\documentclass[conference]{IEEEtran}
\usepackage[compatibility=false]{caption}
\IEEEoverridecommandlockouts

\usepackage{amsmath,amssymb,amsfonts}
\usepackage{algorithmic}
\usepackage{graphicx}
\usepackage{textcomp}
\usepackage{newtxtext}
\usepackage{booktabs}
\usepackage{multirow}
\usepackage{float}
\usepackage{makecell}
\usepackage{subcaption}
\usepackage[square,numbers]{natbib}
\usepackage[table]{xcolor}

\definecolor{citeblue}{rgb}{0.21,0.49,0.74}
\definecolor{autorefcolor}{rgb}{0.95,0.30,0.25}
\usepackage[colorlinks=true, citecolor=citeblue, linkcolor=citeblue]{hyperref}
\usepackage{wrapfig}
\usepackage{xspace}

\newcommand{\ibefore}{\boldsymbol{x}_{t_0}}
\newcommand{\iafter}{\boldsymbol{x}_{t_1}}

\newcommand{\extractor}{\ensuremath{f_{\theta}}}
\newcommand{\cdmodel}{\ensuremath{\mathrm{CD}_{\theta}}}
\newcommand{\levir}{\text{LEVIR-CD}\xspace}
\newcommand{\redd}{\text{REDD-Forest-CD}\xspace}
\newcommand{\ours}{\text{FDC}\xspace}

\def\BibTeX{{\rm B\kern-.05em{\sc i\kern-.025em b}\kern-.08em
T\kern-.1667em\lower.7ex\hbox{E}\kern-.125emX}}
\begin{document}

\title{Fidelity-Diversity-Consistency (FDC): Data Pruning for Remote Sensing Change Detection\\
\thanks{This research is part of AI-LEAF: “AI Institute for Land,
Economy, Agriculture \& Forestry,” and is supported by USDA
National Institute of Food and Agriculture (NIFA) and the
National Science Foundation (NSF) National AI Research
Institutes Competitive Award no. 2023-67021-39829. Project
website: https://cse.umn.edu/aileaf.}
}

\author{
\IEEEauthorblockN{Dongyao Zhu$^{*}$ \quad Ranga Raju Vatsavai}
\IEEEauthorblockA{
Department of Computer Science\\
\textit{North Carolina State University}\\
Raleigh, North Carolina, USA\\
$^{*}$Corresponding author. Email: dzhu6 [at] ncsu [dot] edu
}
}
\maketitle

\begin{abstract}
Despite the success of data pruning (DP) in reducing training data sizes and improving downstream model performance in classification and segmentation tasks, its potential in remote sensing change detection remains unexplored. For the first time, we benchmark six representative DP methods across building- and forest-change datasets, CNN- and transformer-based models, and three pruning budgets, and show that existing baselines yield no reliable advantage over random selection. Notably, even the strongest evaluated baseline, Feature Diversity, is matched or exceeded by $\sim$33\% of randomly sampled subsets. To understand the underlying mechanism, we conduct a systematic regression study over 540 randomly sampled data subsets, characterizing each with four descriptors covering label statistics, image diversity, and feature-space geometry. Random Forest models show that \emph{change distribution fidelity} is the most prominent factor in determining the quality of change detection data subsets, a property absent from the existing pruning literature. Our analyses further show that pixel-wise image diversity and label-feature consistency are secondary factors. We translate these findings into Fidelity-Diversity-Consistency (FDC), a simple two-stage pruning method that shows consistent improvements over existing baselines across change detection benchmarks and backbones, especially at lower pruning ratios. Code is available at \href{https://github.com/ddydyd32/fidelity-diversity-consistency}{https://github.com/ddydyd32/fidelity-diversity-consistency}.
\end{abstract}


\section{Introduction}

The prevailing assumption in deep learning that scaling dataset size consistently yields superior performance has driven collection efforts toward web-scale corpora containing hundreds of millions of examples. Yet this trajectory faces two compounding problems. First, the returns on additional data follow a negative power law: model error scales as $N^{-\alpha}$ relative to dataset size $N$ ~\citep{hestness2017deep}, so each marginal example contributes less than the last while the computational cost of processing each example remains constant. Second, massive uncurated datasets are notoriously noisy, frequently containing mislabeled examples, near-duplicates, and irrelevant content that actively degrades generalization~\citep{gadre2023datacomp, fan2023improving}. Achieving state-of-the-art performance now demands exponentially more data, compute, and energy, which is both financially and environmentally unsustainable.

Data pruning (DP) directly addresses both inefficiencies. By identifying and removing redundant or detrimental examples or selecting informative examples before training, DP methods extract a smaller, high-quality ``coreset'' that enables downstream models to achieve comparable, or occasionally superior, accuracy while drastically reducing training time, energy expenditure, and storage overhead.

\begin{figure}[t]
    \centering
    \includegraphics[width=\linewidth]{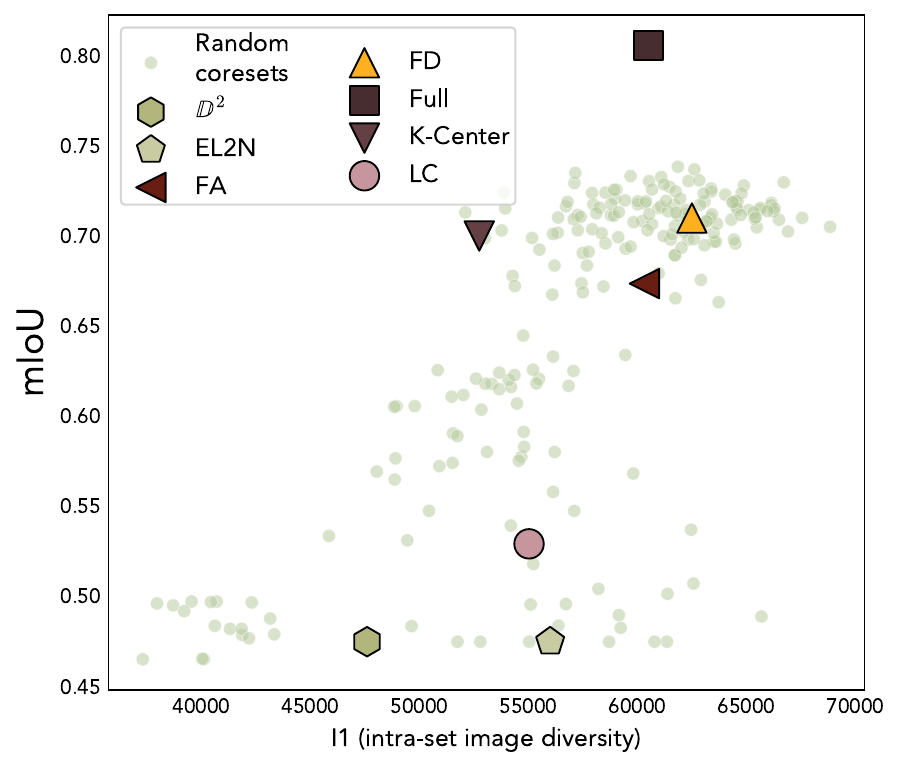}
    \vspace{-12pt}
    \caption{Higher image diversity correlates with better mIoU, still $\sim$33\% of random coresets are competitive with diversity-driven DP methods.}
    \label{fig:teaser}
    \vspace{-10pt}
\end{figure}


\subsection{Datasets in Remote Sensing Based Change Detection}
Remote sensing (RS) data collection is highly scalable through continuous satellite constellations, but acquiring accurate ground-truth annotations remains expensive. For multi-temporal change detection (CD), researchers therefore often rely on ancillary reference products, such as the USGS National Land Cover Database~\citep{dewitz2021national} and the USDA Cropland Data Layer~\citep{boryan2011monitoring}. However, these derivative products contain systematic labeling errors~\citep{townshend1992impact}, spatially generalized boundaries, and temporal mismatches~\citep{liu2024integrated}, introducing substantial label noise into downstream training pipelines.

\subsection{Unique Complexities of Pruning for Change Detection}
Directly applying DP methods to bi- and multi-temporal change detection represents a major gap in the literature due to three compounding challenges unique to this task:
\begin{enumerate}

\item \textbf{Extreme Class Imbalance and Change Distribution Sensitivity:} Semantic changes typically occupy a small fraction of geographic area, leaving most bi-temporal image pairs unchanged~\citep{lei2026remote}. Standard pruning methods may discard scarce ``changed'' samples while retaining redundant ``no-change'' background. Thus, a viable coreset must preserve the dataset's change ratio, or \textit{change distribution fidelity}, which existing DP methods do not explicitly consider.

\item \textbf{Spatial Autocorrelation and Severe Spatial Redundancy:} Following Tobler's First Law of Geography, nearby regions tend to be highly similar, resulting in thousands of correlated patches from homogeneous RS landscapes. Traditional pruning methods struggle to quantify this redundancy~\citep{moser2025coreset}, potentially over-sampling redundant regions while missing localized edge cases. This motivates diversity measures in semantic feature space beyond raw pixel or geographic space.

\item \textbf{Pseudo-Change vs.\ Semantic Change:} Change detection must distinguish true semantic transitions from ``pseudo-changes'' caused by phenology, illumination, or registration errors~\citep{townshend1992impact, liu2024integrated, cheng2024change}. Although pseudo-changes may appear informative under standard diversity metrics, they provide unreliable supervision and can harm training. Thus, pruning should consider \textit{feature-label consistency} alongside feature diversity to remove superficially informative but misleading samples.

\end{enumerate}

Despite these challenges, DP remains under-explored in RS. Recent work studied DP primarily for static land-cover classification and semantic segmentation~\citep{nogueira2026core}, with little attention given to its application in multi-temporal change detection.

In this study, we directly address this gap by adapting and systematically evaluating existing DP methods~\citep{paul2021deep, nogueira2026core} for remote sensing change detection (RS-CD). Despite their success on segmentation and classification tasks, these methods fail to translate effectively to change detection. Across two RS-CD benchmarks~\citep{11402297, Chen2020} and two model backbones~\citep{codegoni2023tinycd, chen2021remote}, we find that none of the evaluated methods consistently outperforms simple random selection (\autoref{tab:bench_tinycd_levir}, \autoref{tab:bench_bit_levir}). In fact, Label Complexity (LC)~\citep{nogueira2026core}, a state-of-the-art method for segmentation data pruning, performs substantially worse than random selection (\autoref{fig:teaser}). Even the strongest method in our RS-CD evaluation, Feature Diversity (FD)~\citep{nogueira2026core}, is matched or exceeded by $\sim$33\% of 260 randomly sampled subsets.

These findings raise the central question of our work: \textit{What properties of a coreset lead to better model performances for remote sensing change detection?} We argue that in addition to diversity, which has been traditionally shown to be important to classification and segmentation tasks, the distribution match between \textit{changes} in the coreset and that in the full dataset is also crucial. To further investigate the importance of coreset from different aspects, we randomly sample a total of 540 random coresets and train the Tiny-CD~\citep{codegoni2023tinycd} model on them. Our regression studies between model performance (mIoU) and four coreset descriptors show that while diversity in pixel space and semantic-feature space correlates to better performance, the change distribution fidelity, approximated by the closeness of a coreset's average change ratio to that of the full dataset, is an indispensable factor to better model performance across different settings, even under extreme pruning ratio at 1\% when diversity is inherently hard to achieve (\autoref{sec:analysis}). 

To this end, we propose a fidelity-diversity-consistency guided two-stage coreset selection method. Motivated by the strong performance of random selection, we start by randomly selecting $pn$ more or $qn$ fewer samples than the target coreset size $n$. We then iteratively prune $pn$ samples or add another $qn$ samples based on each sample's diversity, change distribution fidelity, and label-feature consistency scores. This design strengthens the competitiveness of random selection by removing low-quality samples and adding high-quality ones, without needing to start from scratch and potentially losing overall representativeness. Empirically, our method produces coresets that achieve superior training performance, especially at extreme pruning ratios of 1\% and 5\% (\autoref{sec:benchmark:results}).

Through this work, for the first time, we comprehensively adapt and benchmark existing data pruning methods to remote sensing-based change detection. While diversity is a well-established criterion in selecting data points for other tasks, our results suggest that change distribution fidelity and feature-label consistency are indispensable aspects in improving data pruning for the change detection task.

\section{Related Work}
The concept of a coreset was first formalized by \citet{har2004coresets}, where a small weighted subset of data can provably approximate the full dataset for k-means clustering. Data pruning (DP) extends this concept to deep learning by identifying and retaining only the most informative training samples from a large dataset, thereby reducing computational costs without significantly sacrificing model performance. Existing data pruning methods can be broadly categorized into two paradigms: training-free and training-based approaches.


\noindent \textbf{Training-Free Data Pruning.} Training-free methods estimate sample importance without training on the full dataset. They instead leverage geometric structure (e.g., clustering), intrinsic statistical properties of the data \citep{nogueira2026core, wei2025rs}, pretrained embeddings \citep{sorscher2022beyond, maharana2024, kamal2024effective, dai2025training, griffin2026zero}, or external scoring models \citep{wang2024cliploss}. These approaches are especially appealing when the cost of full-dataset training is prohibitive.

\noindent \textbf{Training-Based Data Pruning.} Training-based methods derive importance scores by observing the behavior of a model trained on the full dataset. These scores capture learning dynamics, such as how quickly and effectively a sample is learned \citep{paul2021deep, he2024large, zhang2024spanning}, how frequently it is forgotten \citep{toneva2018empirical}, or how large its gradient magnitude is \citep{paul2021deep}. Such signals are then used to identify the most informative samples for subsequent model training. While these methods generally achieve stronger performance than training-free approaches, they require either a full preliminary training run or the use of an external proxy model for score estimation, resulting in higher computational costs.

\noindent \textbf{Data Pruning for Remote Sensing.} Data pruning for remote sensing (RS) is an emerging research direction, driven by the rapid growth of large-scale satellite imagery and the increasing cost of annotating RS data. While general-purpose DP methods have demonstrated effectiveness on natural-image benchmarks, the unique challenges of RS data often necessitate domain-specific adaptations, including high spectral dimensionality, severe class imbalance, top-down viewpoints, and noisy pixel-level annotations. For self-supervised learning, a dynamic pruning strategy is proposed in \citep{kerdreux2025efficient} for the Sentinel-1 Wave Mode SAR archive, alternating between clustering-based diversity assessment and coreset selection during pretraining without requiring an external model. For RS semantic segmentation, the first data-centric benchmark is introduced in \citep{nogueira2026core} along with several training-free pruning methods based on image and label quality modeling.

Despite the demonstrated effectiveness of data pruning for classification, segmentation, and self-supervised pretraining in both general computer vision and, more recently, remote sensing, its application to remote sensing change detection (RS-CD) remains unexplored. To the best of our knowledge, this work represents the first systematic study of dataset selection methods for efficient RS-CD model training.

\section{Existing Pruning Criteria Fail to Consistently Outperform Random Selection}
\label{sec:benchmark}

We begin with an empirical study to assess the performance of existing DP methods on CD tasks. Specifically, we evaluate six representative pruning methods across two CD datasets, two CD backbones, and three pruning budgets. The detailed experimental setup is as follows:
 
\label{sec:benchmark:setup}
\subsection{Datasets}

We train and test on 2 remote-sensing change detection benchmarks using their official train/test splits. 

\noindent \textbf{\levir \citep{Chen2020}} is a building change dataset with 637 pairs of $1024 \times 1024$ aerial images (0.5\,m spatial resolution).

\noindent \textbf{\redd \citep{11402297}} is a forest change dataset with 3087 pairs of $512 \times 512$ images from Sentinel-2 \cite{drusch2012sentinel} (10\,m resolution). To ensure consistent label quality, we limit our analysis to data from Bolivia and Brazil with 1054 pairs of examples.

For all datasets, we resize the images to $512 \times 512$ using bilinear interpolation. For pruning ratio settings, we train at 1\%, 5\%, and 10\% of the full training set, corresponding to approximately 5, 23, and 46 samples on \levir.

\subsection{Change Detection Model Backbones}
We evaluate two change detection backbones, Tiny-CD and BIT, representing lightweight CNN-based and CNN–Transformer-based models, respectively, and together covering a range of model capacities and architectural designs.

\noindent \textbf{Tiny-CD} \citep{codegoni2023tinycd} is a lightweight CNN-based siamese U-Net \citep{ronneberger2015u} framework with only $\sim$1.2M parameters. It employs a shared convolutional encoder with a Mix and Attention Mask Block (MAMB) to fuse temporal and semantic information from bi-temporal images while preserving fine spatial details.


\noindent \textbf{BIT} \citep{chen2021remote} combines a ResNet-18 \citep{he2016deep} encoder with a Transformer \citep{vaswani2017attention} module and operates in a compact semantic token space. Bi-temporal image features are first extracted by the shared CNN backbone, then projected into a small set of semantic tokens that interact via multi-head self-attention to capture long-range context. The refined tokens are back-projected onto the spatial feature maps before a convolutional decoder produces the final change mask ($\sim$3.6M parameters).

Both models are trained with a batch size of 8 using a hybrid Binary Cross-Entropy (BCE) and Dice loss \citep{milletari2016v} for robust optimization under class imbalance. Following \citet{nogueira2026core, guo2022deepcore}, we fix the number of training epochs to 100 across all pruned datasets. Following \citet{11402297}, we report F1, Mean Intersection over Union (mIoU), and Cohen's $\kappa$ coefficient as evaluation metrics, averaged over 3 runs. Our implementations are based on Open-CD \citep{opencd}.

\subsection{Baseline Data Pruning Methods}
\label{sec:baselines}
We include two error-based methods (EL2N, $\mathbb{D}^2$), three image-based methods (FA, FD, K-Center), and one label-based method (LC), spanning the main families of data pruning.
Let $\ibefore^i, \iafter^i \in \mathbb{R}^{H \times W \times C}$ be a pair of image inputs at different timestamps and $\boldsymbol{y}^i \in \{0,1\}^{H \times W}$ the pixel-wise change label, for the $i$-th example in the dataset.
Let $\extractor$ denote an image feature encoder and $\cdmodel$ the CD model backbone.
Unless stated otherwise, $\extractor$ is a ResNet-18 \citep{he2016deep} pre-trained for classification on ImageNet-1k \citep{russakovsky2015imagenet} and kept frozen; the same encoder is used for every baseline that needs embeddings, for the descriptors of \autoref{sec:analysis:descriptors}, and for \ours, so that differences between methods reflect the selection criterion rather than the representation.
We note that ImageNet features are learned from single-temporal, object-centric natural images and are therefore an imperfect proxy for multi-temporal RS semantics; they carry no notion of bi-temporal correspondence, and their sensitivity to illumination and seasonal appearance is precisely what makes pseudo-change hard to separate from semantic change.
Our findings should thus be read as characterizing pruning criteria under a standard ImageNet-pretrained encoder, and an RS-specific or bi-temporal pretrained encoder is a natural way to strengthen every feature-based criterion studied here, including ours.

\noindent \textbf{EL2N Score \citep{paul2021deep} (EL2N)}
measures the norm of the error vector between predicted logits and label:
\begin{align*}
\text{s}^i = \mathbb{E}_{\cdmodel} \| p(\cdmodel, \ibefore^i, \iafter^i) - \boldsymbol{y}^i \|_2.
\end{align*}
A higher score indicates a sample still difficult after training.

\noindent \textbf{$\mathbb{D}^2$ Pruning \citep{maharana2024} ($\mathbb{D}^2$)} represents
the dataset as an undirected $k$-NN graph $\mathcal{G}$ and uses message passing to jointly optimize sample difficulty and data diversity. In our adaptation, we use change-weighted EL2N
as the difficulty score to initialize each node in $\mathcal{G}$, and build edge weights using
the RBF kernel over the Euclidean distance ($d$) between difference-image features:
\begin{align*}
    e_{i,j} = \exp\!\left(-\gamma_f \cdot d\!\left(\extractor(|\iafter^i - \ibefore^i|),\, \extractor(|\iafter^j - \ibefore^j|)\right)^2\right),
\end{align*}
where $\extractor$ is a frozen ResNet-18~\citep{he2016deep} and $|\iafter^i - \ibefore^i|$ is the absolute difference image for patch $i$.
Following \citep{maharana2024}, forward message passing aggregates neighbourhood difficulty
scores, and samples are selected greedily with reverse message passing applied after each
selection to suppress redundancy, using a single global budget. 
A higher score indicates a sample that is both difficult and dissimilar from samples already selected.

\noindent \textbf{Feature Activation \citep{nogueira2026core} (FA)} is an image-based baseline that ranks training examples using statistics derived from image embeddings produced by a pre-trained ResNet-18 \citep{he2016deep} network $\extractor$. For each example $i$, the embedding is computed as $\extractor(|\iafter^i - \ibefore^i|) \in \mathbb{R}^{512}$, and a score $s^{i}$ is derived from its mean $\mu_i$ and standard deviation $\sigma_i$, both normalized to $(0, 1]$:
\begin{align*}
    s^{i} = 1 - \left[ \frac{ \gamma_i - \min\limits_{j}[\gamma_j]}{\max\limits_{j}[\gamma_j]-\min\limits_{j}[\gamma_j]}\right], \quad \gamma_i = -(1-\mu_i) \cdot \log(\sigma_i).
\end{align*}
A higher score indicates greater diversity and less noise.

\noindent \textbf{Feature Density \citep{nogueira2026core} (FD)} extracts feature embeddings in the same way as in FA for each example, but uses them to perform diversity-aware selection via K-Means clustering. The embeddings are grouped into $K$ clusters, where $K$ equals the target number of samples to retain after pruning. Samples are then selected in a round-robin fashion across clusters, prioritizing those closest to their cluster centroid, such that examples farther from any centroid receive a lower score $s^i$.

\noindent \textbf{K-Center Greedy \citep{guo2022deepcore} (K-Center)} extracts embeddings the same as in FA and FD, but selects samples by solving a minimax facility location problem \citep{farahani2009facility}. We use the greedy $\mathcal{O}(N \times k)$ approximation \citep{sener2017active} that iteratively selects the sample with the largest minimum Euclidean distance to the current selected set in feature space. Examples selected earlier are more isolated from others, reflecting greater spatial diversity in the embedding space, and are therefore assigned higher scores.

\noindent \textbf{Label Complexity \citep{nogueira2026core} (LC)} is a label-based method that scores each example based on the entropy of its change mask $\boldsymbol{y}^i$. For each example $i$, the score is computed as:
\begin{equation}
    s^i = H(\boldsymbol{y}^i) = - \sum_{c=1}^{C} p_{i,c} \log_C (p_{i,c}),
\end{equation}
where $C$ is the number of classes and $p_{i,c}$ is the proportion of pixels of class $c$ in $\boldsymbol{y}^i$. Examples with higher scores exhibit a more uniform class distribution, reflecting greater label complexity and potentially richer supervision signal.

\begin{figure}[t]
    \centering
    \includegraphics[width=\linewidth]{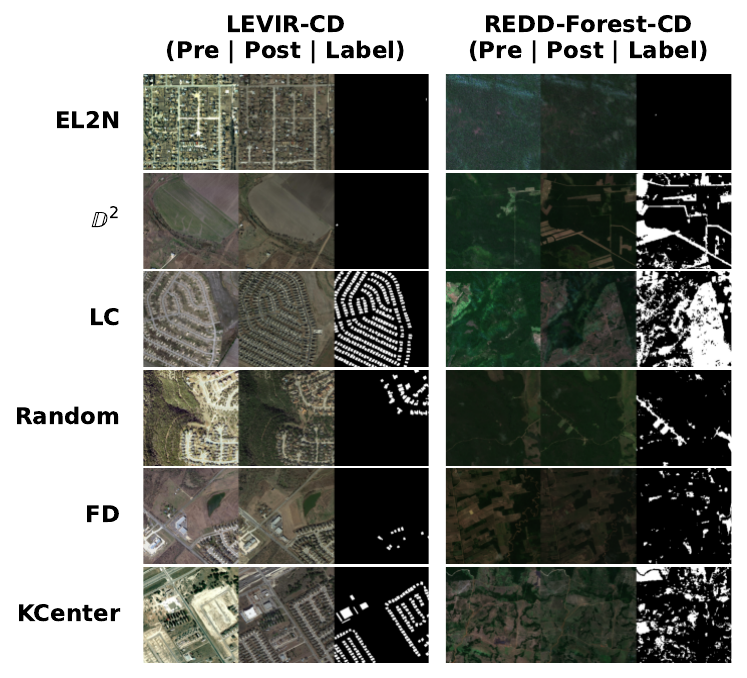}
    \caption{The first CD examples chosen by different pruning methods. Loss-based EL2N and $\mathbb{D}^2$ picks examples with minimal or ambiguous changes. Label-based LC picks an example with substantial changes. Diversity-based K-Center and FD pick first examples similar to random selection.}
    \label{fig:qualitative}

    \vspace{-8pt}
\end{figure}


\subsection{Benchmark Results} 
\label{sec:benchmark:results}

\autoref{tab:bench_tinycd_levir}, \autoref{tab:bench_bit_levir} report results for the Tiny-CD and BIT backbones on \levir and \redd.

\noindent \textbf{Existing baselines yield no reliable advantage over random selection.} 
Among all adapted baselines, only feature-geometry methods are competitive with random selection, and only sporadically. FD achieves the highest score among baselines at 5\% on \levir (F$_1$ of $64.5$ vs.\ random $62.6$), while K-Center dominates at the 1\% budget on both datasets (F$_1$ of $44.2$ on \levir and $51.8$ on \redd, vs.\ random at $23.8$ on both datasets).
This suggests that at very small budgets, maximizing geometric spread in feature space is the most reliable individual heuristic: 
without deliberate coverage, random sampling risks leaving catastrophic holes in feature-space coverage.
Yet no single method dominates across all budget--dataset combinations. K-Center beats random on \levir at 10\% (F$_1$ $69.5$ vs.\ $68.1$) but underperforms on \redd ($55.0$ vs.\ $58.8$); at 5\%, this pattern reverses: K-Center loses to random on \levir while FD surpasses it, yet FD in turn falls below \ours-P on the same split.
The inconsistency stands in sharp contrast to image classification~\citep{paul2021deep,sorscher2022beyond} and semantic segmentation~\citep{dai2025training}, where well-designed pruning criteria yield reliable gains.
We attribute the difficulty in RS-CD to the joint dependence of change labels on \emph{pairs} of images: a sample that is geometrically or semantically representative in single-image feature space may still fail to capture the conditional distribution of change patterns, making naive quality proxies unreliable.

\noindent \textbf{Loss-based methods fail catastrophically.}
EL2N collapses to trivial predictions ($F_1 = 0$) at every pruning ratio on \levir and at $\leq$5\% on \redd.
We attribute this to a fundamental mismatch between what ``hardness'' indicates in classification versus change detection.
In classification, high-loss samples lie near decision boundaries and carry discriminative information.
In CD, high-loss patches are dominated by visually complex but semantically unchanged regions (e.g.\ seasonal foliage differences, shadow shifts, atmospheric variation) that produce large pixel-level prediction errors despite containing no labelled structural change.
EL2N thus preferentially selects these \emph{visually distracting} patches as illustrated in \autoref{fig:qualitative}, yielding a coreset with little or no positive change supervision, causing the model to predict the majority ``no-change'' class everywhere.
$\mathbb{D}^2$, which reweights EL2N by $k$-NN diversity, partially rescues performance on \redd (F$_1$ $53.9$ at 5\%), where baseline change ratios are higher and diversity reweighting distributes the pool more evenly, but still fails on \levir.
This demonstrates that the problem is not merely class imbalance but a decorrelation between prediction difficulty and supervision quality intrinsic to change detection.

\noindent \textbf{Label-complexity selection is harmful.} 
\label{sec:lc_fails}
LC underperforms random in nearly every setting, often dramatically (F$_1$ $36.9$ vs.\ $62.6$ at 5\% on \levir; $34.0$ vs.\ $53.9$ on \redd).
By optimizing solely for per-sample mask entropy, LC selects samples from a narrow band of frequent change ratios as shown in \autoref{fig:qualitative}, destroying the distributional coverage of the resulting subset.
This illustrates a failure mode specific to set-level selection: a locally informative criterion (each sample has high label entropy) can be globally destructive (the subset has low distributional diversity).

\noindent \textbf{Model architecture has a limited moderating effect.}
Comparing the CNN-based Tiny-CD with the transformer-based BIT, the overall ranking of methods is broadly preserved: loss-based methods fail on both, geometric methods are competitive at low ratios, and LC consistently hurts.
BIT exhibits a slightly sharper degradation curve as the budget shrinks, likely because its stronger backbone requires more data, but this does not alter the overall ranking of the methods.
This implies that the difficulty of data pruning for RS-CD is rooted in the dataset structure rather than model architecture.

Together, these findings motivate a closer examination of the random subsets themselves to identify the intrinsic properties that actually drive change detection performance.

\section{What Makes a Good Change Detection Coreset?}
\label{sec:analysis}

To answer this question and better understand why random selection can be competitive, we perform a controlled regression analysis on random data coresets, using a suite of training-free descriptors to characterize each subset and identify which properties correlate with downstream model performance.

\subsection{Generating (subset, performance) pairs} 
\label{sec:analysis:design}
We generate random coresets at both 5\% and 1\% ratios, where random often beats other methods, and the training cost is acceptable. For each ratio, 100 coresets are sampled uniformly and another 160 (\levir) or 180 (\redd) from stratified change-ratio bins to ensure coverage of the change-ratio distribution, which is noted to be long-tailed in \citep{lei2026remote}. Each coreset is used to train Tiny-CD for 100 epochs, and mIoU is recorded after averaging over three random seeds, resulting in a descriptor matrix $\mathbf{D} \in \mathbb{R}^{N \times d}$
and performance vector $\mathbf{P} \in \mathbb{R}^{N\times1}$.

\subsection{Descriptor Design} 
\label{sec:analysis:descriptors}
We design 4 descriptors per subset from three signal sources: images, change labels, and features computed by \extractor. All descriptors are computed directly from raw images and labels; no CD model training is required. 
We follow the notations in \autoref{sec:baselines}.

\noindent \textbf{Label-space descriptor.} 

\noindent \textbf{L1} \emph{Change Distribution Fidelity}: approximated with ratio deviation $\delta{r} = 1 - |(\frac{1}{k} \sum_{i=1}^{k} \frac{\sum \boldsymbol{y}^i}{HW} - r_{\text{full}})|$, where $r_{\text{full}} = \frac{1}{N} \sum_{j=1}^{N} \frac{\sum \boldsymbol{y}^j}{HW}$ is the change ratio of the full dataset. Measures the deviation between the average change ratio of the selected subset and that of the full dataset.



\noindent \textbf{Image-space descriptor.} 



\noindent \textbf{I1} \emph{Intra-Set Image Diversity}: mean pairwise $\ell_2$ distance among pre-images $\{\ibefore^i\}_{i=1}^{k}$. Measures the visual diversity of the subset, with larger values indicating lower pixel-level redundancy.

\noindent \textbf{Feature-space descriptors.} 

\noindent \textbf{F1} \emph{Intra-Set Feature Diversity}: 
mean pairwise cosine distance among concatenated features $\{[\extractor(\ibefore^i), \extractor(\iafter^i)]\}_{i=1}^{k}$. Measures the semantic diversity of the subset, with larger values indicating lower feature-level redundancy.



\noindent \textbf{F2} \emph{Label-Feature Consistency}: Pearson correlation between
per-sample feature similarity,
$\cos(\extractor(\ibefore^i), \extractor(\iafter^i))$,
and change ratio $r_i$ across the subset.
Measures how well semantic appearance changes correspond to labeled changes.






\begin{table}[t]
\centering
\caption{Univariate $R^2$ and random forest (RF) permutation importance of descriptors on random subsets vs. mIoU.}
\label{tab:descriptors}
\setlength{\tabcolsep}{3pt}
\renewcommand{\arraystretch}{1.0}
\resizebox{0.9\linewidth}{!}{%
\begin{tabular}{l l cc | cc}
\toprule
\multirow{2}{*}{\textbf{Dataset}} & \multirow{2}{*}{\textbf{Desc.}} 
& \multicolumn{2}{c}{5\% \textbf{Pruning}}
& \multicolumn{2}{c}{1\% \textbf{Pruning}} \\
\cmidrule(lr){3-4}\cmidrule(lr){5-6}
& & $R^2$ & $\text{RF}$ & $R^2$ & $\text{RF}$ \\
\midrule
\multirow{4}{*}{\levir}
& I1 & 0.654 & 0.479 & 0.077 & 0.069 \\
& L1 & 0.413 & 0.750 & 0.000 & 1.727 \\
& F1 & 0.441 & 0.012 & 0.027 & 0.072 \\
& F2 & 0.119 & 0.003 & 0.036 & 0.063 \\
\midrule
\multirow{4}{*}{\redd}
& I1 & 0.711 & 0.005 & 0.528 & 0.023 \\
& L1 & 0.889 & 1.781 & 0.823 & 1.722 \\
& F1 & 0.655 & 0.015 & 0.528 & 0.020 \\
& F2 & 0.001 & 0.007 & 0.005 & 0.025 \\

\bottomrule
\end{tabular}%
}
\vspace{-8pt}
\end{table}

\subsection{Analysis on coreset descriptors vs.\ model performance} 

We apply two complementary analyses: (1.) univariate linear regression ranked by $R^2$, (2.) random forest (RF) with permutation importance as a nonlinear robustness check.
The same pipeline is run at both 5\% and 1\% pruning regimes on both datasets.
The regression results are reported in \autoref{tab:descriptors}.

The most striking finding is that change distribution fidelity (L1) is the single most important predictor across both datasets and pruning regimes, while intra-set diversity (I1) provides a complementary, dataset-dependent boost that becomes more prominent at moderate budgets.

\noindent \textbf{Change distribution fidelity dominates, with diversity as a dataset-dependent complement on \levir.}
At 5\% pruning on \levir, the two analysis views offer complementary perspectives.
RF permutation importance ranks L1 highest ($0.750$), confirming that change-ratio matching is the most influential factor even in this visually heterogeneous dataset.
Meanwhile, I1 leads in univariate $R^2$ ($0.654$ vs.\ $0.413$ for L1), showing that intra-set image diversity itself explains more of the linear variance in mIoU.
Together, they reveal that L1 is the primary driver of performance, while I1 provides a significant supplementary signal to improve model quality.
F1 (intra-set feature diversity, $R^2 = 0.441$) further supports the value of semantic coverage as a secondary criterion.

\noindent \textbf{Change distribution fidelity is the near-sufficient predictor on \redd.}
\label{sec:analysis_redd}
On \redd, L1 is unambiguously the most important factor at both 5\% and 1\% pruning, accounting for $R^2 = 0.889$ (5\%) and $0.823$ (1\%) of mIoU variance, respectively, and dominating RF importance ($1.781$ and $1.722$).
That is, subsets whose change-ratio distribution closely matches the full dataset are strongly preferred.
This result is consistent with \redd's structure: the dataset contains mostly unchanged forest images with low visual diversity (trees and mountains), thus diversity-based selection helps little when most images look similar. 
Instead, the balance of changes to include in the coreset is more critical: too few changed images and the model lacks positive supervision; too many and it overly predicts change.
This indicates that a hard constraint on distributional fidelity (ensuring L1 stays close to the population mean) is required.

\noindent \textbf{At extreme budgets, the role of diversity diverges across datasets.}
\label{sec:analysis_transition}
At 1\% pruning, L1 dominates RF importance on both datasets (\levir: $1.727$; \redd: $1.722$), confirming that change-ratio fidelity is a necessary condition at extreme budgets where even a single missing change-class example can eliminate positive supervision.
However, the univariate $R^2$ tells a more nuanced story.
On \levir, all individual $R^2$ values collapse (I1: $0.077$, L1 drops to zero), suggesting that with only 4-5 examples, no single descriptor linearly predicts performance. 
On \redd, I1's univariate $R^2$ remains substantial ($0.528$), indicating that even within a visually homogeneous dataset, diversity begins to differentiate coresets once distributional fidelity is controlled for.
This budget-dependent transition reveals a priority hierarchy: distributional fidelity is a binding constraint at extreme budgets, while diversity provides additional gains once the distributional requirement is met, and its relative contribution increases as the budget shrinks.


\begin{figure}[t]
    \centering
    \includegraphics[width=0.9\linewidth]{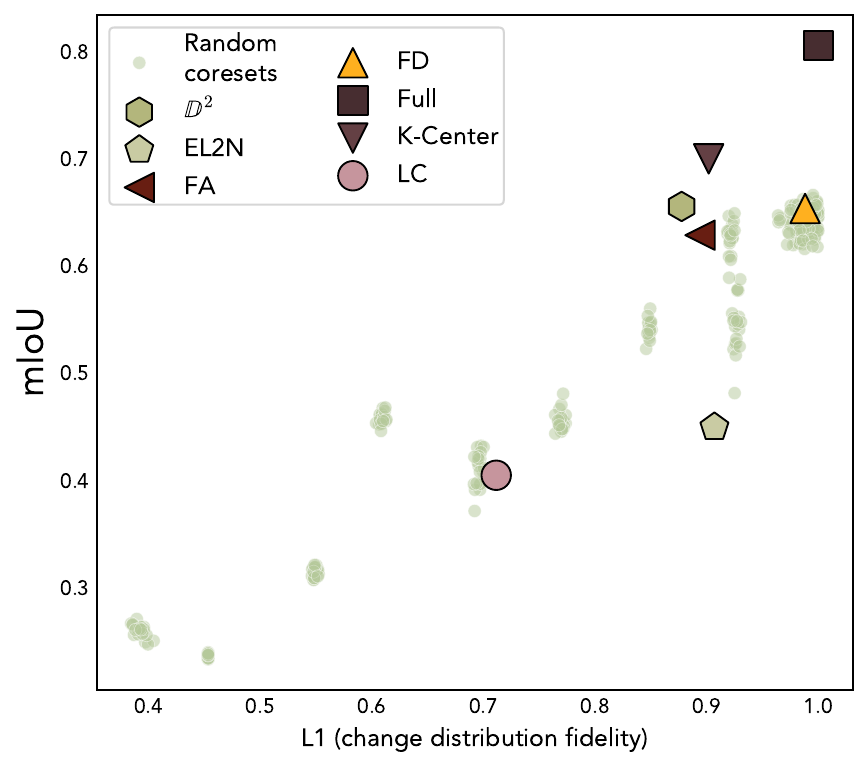}
    \vspace{-6pt}
    \caption{Higher change distribution fidelity (L1) correlates to better mIoU.}
    \vspace{-12pt}
    \label{fig:x2}
\end{figure}

\noindent \textbf{Implications for pruning criterion design.}
\label{sec:analysis_implications}

The descriptor analysis reveals that L1 is the universally dominant predictor, but effective coreset construction requires addressing a second, dataset-dependent axis:

\begin{itemize}
    \item \textbf{Diversity-sensitive regime} (\levir-like): L1 is necessary, but I1 provides a large complementary boost. Scene-type coverage determines model quality beyond what distributional matching alone achieves.
    \item \textbf{Distribution-dominated regime} (\redd-like): L1 is near-sufficient. Visual diversity is low by nature, so distributional fidelity explains nearly all performance variance.
\end{itemize}

At lower pruning ratios, distributional fidelity becomes a harder constraint, while the marginal value of diversity increases, particularly on datasets with greater visual heterogeneity.
Crucially, the best-performing random subsets succeed on \emph{both} axes simultaneously: they are diverse (\autoref{fig:teaser}) \emph{and} representative in change distribution (\autoref{fig:x2}).
Existing methods address only one axis (LC targets change-ratio balance but ignore diversity; FD and K-Center target diversity but ignore change-ratio matching), explaining their inconsistency.
F2 provides a third, complementary signal: label-grounded quality, which can further discriminate among otherwise comparable candidates.
These observations directly motivate the design of our proposed method in \autoref{sec:our_method}.

\section{Method: Fidelity, Diversity, Consistency}
\label{sec:our_method}

We propose Fidelity-Diversity-Consistency (\textbf{\ours}), a two-phase coreset method that directly operationalizes previous findings.
A candidate pool is first initialized with the competitive random selection, then iteratively refined using descriptor-based scoring. We design two complementary variants:

\noindent \textbf{\ours-P (Prune):} oversample the random candidate pool by $p\%$ samples and iteratively remove low-quality samples.

\noindent \textbf{\ours-G (Grow):} start with a random subset with $g\%$ of target size and iteratively add high-quality samples.

Based on our previous findings in \autoref{sec:benchmark:results}, 
at higher pruning ratios $r = $ 10\% and 5\% where random is competitive, given a target coreset size $n$, we initialize \ours-P with $pn$ additional samples, then remove $pn$ low-quality samples. At a pruning ratio of 1\%, where random starts to degrade, we initialize \ours-G with $gn$ fewer samples, then add another $gn$ high-quality samples. We choose $p = 20\%$ and $g = 40\%$ by default, detailed in \autoref{sec:ablation_pg}.
Both method variants use I1, L1, F1, F2 to score each sample, all normalized to $[0,1]$ and linearly combined as:

\begin{equation}
\text{score} = s_i^{\text{\ours}} = \alpha \cdot I_1^i + \beta \cdot L_1^i + \gamma \cdot F_1^i + \delta \cdot F_2^i
\label{eq:eq2}
\end{equation}

where $\alpha$, $\beta$, $\gamma$, and $\delta$ control the relative importance of each term. 
Note that the score in \autoref{eq:eq2} is defined at the set level. During iterative pruning or growing, we evaluate each candidate by the marginal utility of the resulting coreset. Specifically, we recompute the FDC score after each candidate addition or removal and select the sample that yields the largest improvement over the current coreset score.

Following our previous analyses in \autoref{sec:analysis_implications}, 
for the diversity-sensitive regime (\levir), we ensure both diversity and change distribution fidelity with $\alpha=1.0$, $\beta=1.0$, $\gamma=0.1$, $\delta=0.1$.
For the distribution-dominated regime (\redd), we ensure change distribution fidelity first with $\alpha=0.1$, $\beta=1.0$, $\gamma=0.1$, $\delta=0.1$. The values $1.0$ and $0.1$ are chosen to reflect the relative importance of the corresponding descriptor in the regression analysis of \autoref{sec:analysis_implications} for a dataset's regime, while keeping every weight in a common, easily interpretable numerical range. Delicate control over each descriptor's contribution, e.g., scaled by regression effect size, is a promising direction for future work.

\section{Experiments}
\label{sec:experiments}

\subsection{Main Results}
\label{sec:main_results}

\autoref{tab:bench_tinycd_levir} and \autoref{tab:bench_bit_levir} compare performances of \ours with other methods.

\noindent \textbf{\ours-P consistently improves over random selection at moderate budgets.}
At 10\% and 5\% pruning, where random selection is already competitive (\autoref{sec:benchmark:results}), \ours-P refines a slightly oversampled random pool by removing low-quality samples.
On Tiny-CD, \ours-P matches or exceeds random selection at every setting. At 10\% on \levir, it improves F1 from $68.1$ to $68.6$ and mIoU from $74.2$ to $74.5$; at 5\%, it achieves the best performance on \levir (F1 $64.7$, mIoU $72.3$). \ours-P also performs the best on \redd at 5\% among all methods (F1 $56.4$, mIoU $65.9$), surpassing both FD ($55.0$, $65.0$) and random selection ($53.9$, $64.5$).
A similar pattern holds for BIT. At 5\% on \levir, \ours-P is the best-performing method (F1 $61.8$, mIoU $70.1$), outperforming FD ($60.8$, $69.6$) and random selection ($59.7$, $68.6$). At 10\%, \ours-P matches FD on \levir (both mIoU $70.6$) and remains close on \redd ($63.1$ vs.\ $63.4$ mIoU).
Unlike baselines that perform strongly on one dataset but weakly on the other (e.g., K-Center excels on \levir at 10\% but underperforms random selection on \redd), \ours-P delivers consistent improvements across both datasets, suggesting that jointly optimizing diversity, fidelity, and consistency avoids the single-axis failure modes of existing methods.

\begin{table}[t]
\centering
\footnotesize
\setlength{\tabcolsep}{3pt}
\caption{Benchmark results (\%, except for $\kappa$) with Tiny-CD on \levir and \redd. Best results per pruning ratio and dataset are in \textbf{bold}. \ours is competitive with all baselines and consistently stable relative to random selection.}\label{tab:bench_tinycd_levir}
\resizebox{\linewidth}{!}{%
\begin{tabular}{cl ccc | ccc}
\toprule
& & \multicolumn{3}{c}{\levir} &
    \multicolumn{3}{c}{\redd} \\
\cmidrule(lr){3-5} \cmidrule(lr){6-8}
Ratio & Pruner & F1 & mIoU & $\kappa$ & F1 & mIoU & $\kappa$ \\

\midrule
100\% & Full                         & 76.9$_{\pm0.7}$ & 80.0$_{\pm0.5}$ & 0.76$_{\pm0.01}$ & 58.3$_{\pm1.4}$ & 67.2$_{\pm0.7}$ & 0.55$_{\pm0.01}$ \\
\midrule
\multirow{8}{*}{10\%}
  & Random                           & 68.1$_{\pm0.8}$ & 74.2$_{\pm0.4}$ & 0.67$_{\pm0.01}$ & 58.8$_{\pm1.3}$ & 66.9$_{\pm0.7}$ & 0.55$_{\pm0.01}$ \\
  & EL2N                             & 7.2$_{\pm10.1}$ & 49.5$_{\pm2.9}$ & 0.07$_{\pm0.09}$ & 0.9$_{\pm0.6}$ & 45.2$_{\pm0.2}$ & 0.01$_{\pm0.01}$ \\
  & $\mathbb{D}^2$                   & 24.6$_{\pm18.5}$ & 55.3$_{\pm6.1}$ & 0.23$_{\pm0.18}$ & 55.9$_{\pm1.3}$ & 65.6$_{\pm0.6}$ & 0.52$_{\pm0.01}$ \\
  & FA                               & 65.4$_{\pm1.6}$ & 72.2$_{\pm1.1}$ & 0.63$_{\pm0.02}$ & 52.4$_{\pm1.3}$ & 64.0$_{\pm0.6}$ & 0.49$_{\pm0.01}$ \\
  & FD                               & 68.9$_{\pm1.5}$ & 74.7$_{\pm0.9}$ & 0.67$_{\pm0.02}$ & \textbf{58.4$_{\pm1.0}$} & \textbf{66.8$_{\pm0.6}$} & \textbf{0.55$_{\pm0.01}$} \\
  & K-Center                         & \textbf{69.5$_{\pm1.0}$} & \textbf{74.8$_{\pm0.7}$} & \textbf{0.68$_{\pm0.01}$} & 55.0$_{\pm0.6}$ & 64.8$_{\pm0.3}$ & 0.51$_{\pm0.01}$ \\
  & LC                               & 58.6$_{\pm0.8}$ & 67.5$_{\pm0.5}$ & 0.56$_{\pm0.01}$ & 38.1$_{\pm2.0}$ & 47.9$_{\pm1.8}$ & 0.27$_{\pm0.03}$ \\
\rowcolor[HTML]{EEB3AA}
  & \ours-P (Ours)                     & 68.6$_{\pm1.3}$ & 74.5$_{\pm0.8}$ & \textbf{0.67}$_{\pm0.01}$ & 56.0$_{\pm1.5}$ & 65.8$_{\pm0.8}$ & 0.53$_{\pm0.01}$ \\
  \midrule
\multirow{8}{*}{5\%}
  & Random                           & 62.6$_{\pm2.8}$ & 71.1$_{\pm1.5}$ & 0.61$_{\pm0.03}$ & 53.9$_{\pm1.2}$ & 64.5$_{\pm0.5}$ & 0.50$_{\pm0.01}$ \\
  & EL2N                             & 0.0$_{\pm0.0}$ & 47.5$_{\pm0.0}$ & 0.00$_{\pm0.00}$ & 0.2$_{\pm0.3}$ & 45.0$_{\pm0.0}$ & -0.00$_{\pm0.00}$ \\
  & $\mathbb{D}^2$                   & 0.0$_{\pm0.0}$ & 47.4$_{\pm0.0}$ & 0.00$_{\pm0.00}$ & 53.9$_{\pm2.1}$ & 64.3$_{\pm1.0}$ & 0.50$_{\pm0.02}$ \\
  & FA                               & 57.2$_{\pm0.3}$ & 67.1$_{\pm0.2}$ & 0.54$_{\pm0.00}$ & 56.2$_{\pm2.2}$ & 65.1$_{\pm1.6}$ & 0.52$_{\pm0.03}$ \\
  & FD                               & 64.5$_{\pm1.3}$ & 72.1$_{\pm0.8}$ & \textbf{0.63$_{\pm0.01}$} & 55.0$_{\pm0.5}$ & 65.0$_{\pm0.2}$ & 0.51$_{\pm0.00}$ \\
  & K-Center                         & 60.2$_{\pm1.3}$ & 68.9$_{\pm0.8}$ & 0.58$_{\pm0.01}$ & 56.2$_{\pm1.2}$ & 65.3$_{\pm0.8}$ & 0.52$_{\pm0.01}$ \\
  & LC                               & 36.9$_{\pm0.4}$ & 53.3$_{\pm0.3}$ & 0.31$_{\pm0.00}$ & 34.0$_{\pm1.9}$ & 43.7$_{\pm2.6}$ & 0.22$_{\pm0.03}$ \\
\rowcolor[HTML]{EEB3AA}
  & \ours-P                          & \textbf{64.7}$_{\pm1.8}$ & \textbf{72.3}$_{\pm1.0}$ & \textbf{0.63}$_{\pm0.02}$ & \textbf{56.4}$_{\pm2.2}$ & \textbf{65.9}$_{\pm1.1}$ & \textbf{0.53}$_{\pm0.02}$ \\
  \midrule
\multirow{8}{*}{1\%}
  & Random                           & 23.8$_{\pm16.1}$ & 54.5$_{\pm4.8}$ & 0.22$_{\pm0.15}$ & 23.8$_{\pm8.9}$ & 52.1$_{\pm3.0}$ & 0.21$_{\pm0.08}$ \\
  & EL2N                             & 0.0$_{\pm0.0}$ & 47.4$_{\pm0.0}$ & 0.00$_{\pm0.00}$ & 0.0$_{\pm0.0}$ & 45.0$_{\pm0.0}$ & 0.00$_{\pm0.00}$ \\
  & $\mathbb{D}^2$                   & 0.0$_{\pm0.0}$ & 47.4$_{\pm0.0}$ & 0.00$_{\pm0.00}$ & 48.3$_{\pm2.0}$ & 59.4$_{\pm1.7}$ & 0.42$_{\pm0.03}$ \\
  & FA                               & 40.0$_{\pm0.6}$ & 57.9$_{\pm0.4}$ & 0.36$_{\pm0.01}$ & 34.0$_{\pm1.1}$ & 45.9$_{\pm2.1}$ & 0.22$_{\pm0.02}$ \\
  & FD                               & 39.6$_{\pm5.1}$ & 59.6$_{\pm1.8}$ & 0.37$_{\pm0.05}$ & 47.4$_{\pm3.5}$ & 61.2$_{\pm1.6}$ & 0.43$_{\pm0.03}$ \\
  & K-Center                         & \textbf{44.2$_{\pm1.1}$} & 60.4$_{\pm0.7}$ & \textbf{0.40$_{\pm0.01}$} & 51.8$_{\pm1.6}$ & 62.4$_{\pm0.7}$ & 0.47$_{\pm0.02}$ \\
  & LC                               & 23.4$_{\pm0.7}$ & 43.6$_{\pm0.7}$ & 0.16$_{\pm0.01}$ & 28.9$_{\pm1.4}$ & 38.5$_{\pm2.2}$ & 0.15$_{\pm0.02}$ \\
\rowcolor[HTML]{EEB3AA}
  & \ours-G (Ours)                     & 41.1$_{\pm2.6}$ & \textbf{60.5}$_{\pm1.1}$ & 0.39$_{\pm0.03}$ & \textbf{53.2$_{\pm1.7}$} & \textbf{63.7$_{\pm0.5}$} & \textbf{0.49$_{\pm0.01}$} \\  
\bottomrule
\end{tabular}
}
\end{table}

\begin{table}[t]
\centering
\footnotesize
\setlength{\tabcolsep}{3pt}
\caption{Benchmark results (\%, except for $\kappa$) of BIT backbone on \levir and \redd. Best results per pruning ratio and dataset are in \textbf{bold}. \ours is competitive with all baselines and consistently stable against random selection.}\label{tab:bench_bit_levir}
\resizebox{\linewidth}{!}{%
\begin{tabular}{cl ccc | ccc}
\toprule
& & \multicolumn{3}{c}{\levir} &
    \multicolumn{3}{c}{\redd} \\
\cmidrule(lr){3-5} \cmidrule(lr){6-8}
Ratio & Pruner & F1 & mIoU & $\kappa$ & F1 & mIoU & $\kappa$ \\
\midrule
100\% & Full                         & 72.0$_{\pm0.7}$ & 76.5$_{\pm0.4}$ & 0.70$_{\pm0.01}$ & 61.1$_{\pm0.5}$ & 68.2$_{\pm0.3}$ & 0.57$_{\pm0.00}$ \\
\midrule
\multirow{8}{*}{10\%}
  & Random                           & 61.8$_{\pm1.6}$ & 70.1$_{\pm0.9}$ & 0.60$_{\pm0.02}$ & 50.6$_{\pm1.0}$ & 62.0$_{\pm0.6}$ & 0.45$_{\pm0.01}$ \\
  & EL2N                             & 0.0$_{\pm0.0}$ & 47.4$_{\pm0.0}$ & 0.00$_{\pm0.00}$ & 0.0$_{\pm0.0}$ & 45.0$_{\pm0.0}$ & 0.00$_{\pm0.00}$ \\
  & $\mathbb{D}^2$                   & 28.4$_{\pm20.1}$ & 55.8$_{\pm5.9}$ & 0.26$_{\pm0.18}$ & 51.9$_{\pm1.3}$ & 62.3$_{\pm0.9}$ & 0.46$_{\pm0.02}$ \\
  & FA                               & 56.6$_{\pm5.7}$ & 67.5$_{\pm2.8}$ & 0.54$_{\pm0.06}$ & 49.7$_{\pm2.7}$ & 60.5$_{\pm1.8}$ & 0.43$_{\pm0.03}$ \\
  & FD                               & 62.4$_{\pm1.5}$ & 70.6$_{\pm0.9}$ & 0.60$_{\pm0.02}$ & \textbf{53.3$_{\pm2.5}$} & \textbf{63.4$_{\pm1.3}$} & \textbf{0.48$_{\pm0.03}$} \\
  & K-Center                         & \textbf{64.6$_{\pm0.9}$} & \textbf{71.5$_{\pm0.6}$} & \textbf{0.62$_{\pm0.01}$} & 51.2$_{\pm0.6}$ & 61.3$_{\pm0.5}$ & 0.45$_{\pm0.01}$ \\
  & LC                               & 62.1$_{\pm2.5}$ & 70.0$_{\pm1.6}$ & 0.60$_{\pm0.03}$ & 33.5$_{\pm1.7}$ & 41.6$_{\pm2.1}$ & 0.21$_{\pm0.02}$ \\
\rowcolor[HTML]{EEB3AA}
  & \ours-P (Ours)                     & 62.4$_{\pm2.0}$ & 70.6$_{\pm1.2}$ & 0.60$_{\pm0.02}$ & 52.5$_{\pm2.2}$ & 63.1$_{\pm1.3}$ & \textbf{0.48$_{\pm0.02}$} \\
\midrule
\multirow{8}{*}{5\%}
  & Random                           & 59.7$_{\pm1.3}$ & 68.6$_{\pm0.8}$ & 0.57$_{\pm0.01}$ & 47.7$_{\pm4.9}$ & 60.6$_{\pm2.3}$ & 0.42$_{\pm0.05}$ \\
  & EL2N                             & 0.0$_{\pm0.0}$ & 47.4$_{\pm0.0}$ & 0.00$_{\pm0.00}$ & 0.0$_{\pm0.0}$ & 45.0$_{\pm0.0}$ & 0.00$_{\pm0.00}$ \\
  & $\mathbb{D}^2$                   & 0.1$_{\pm0.2}$ & 47.5$_{\pm0.0}$ & 0.00$_{\pm0.00}$ & 47.9$_{\pm1.5}$ & 59.9$_{\pm1.0}$ & 0.42$_{\pm0.02}$ \\
  & FA                               & 56.2$_{\pm2.6}$ & 66.7$_{\pm1.5}$ & 0.53$_{\pm0.03}$ & 46.7$_{\pm1.4}$ & 57.5$_{\pm1.2}$ & 0.39$_{\pm0.02}$ \\
  & FD                               & 60.8$_{\pm1.3}$ & 69.6$_{\pm0.8}$ & \textbf{0.59$_{\pm0.01}$} & \textbf{51.3$_{\pm1.5}$} & \textbf{61.9$_{\pm0.8}$} & \textbf{0.46$_{\pm0.02}$} \\
  & K-Center                         & 59.0$_{\pm2.2}$ & 68.3$_{\pm1.3}$ & 0.56$_{\pm0.02}$ & 45.7$_{\pm4.5}$ & 57.4$_{\pm2.9}$ & 0.38$_{\pm0.05}$ \\
  & LC                               & 49.6$_{\pm1.0}$ & 62.6$_{\pm0.9}$ & 0.46$_{\pm0.01}$ & 26.7$_{\pm3.3}$ & 29.9$_{\pm6.6}$ & 0.12$_{\pm0.05}$ \\
\rowcolor[HTML]{EEB3AA}
  & \ours-P (Ours)                     & \textbf{61.8$_{\pm1.9}$} & \textbf{70.1$_{\pm1.2}$} & \textbf{0.59$_{\pm0.02}$} & 49.1$_{\pm4.5}$ & 61.3$_{\pm2.1}$ & 0.44$_{\pm0.05}$ \\
  \midrule
\multirow{8}{*}{1\%}
  & Random                           & 13.9$_{\pm19.2}$ & 51.7$_{\pm5.9}$ & 0.13$_{\pm0.18}$ & 6.3$_{\pm4.3}$ & 46.2$_{\pm0.9}$ & 0.04$_{\pm0.03}$ \\
  & EL2N                             & 0.0$_{\pm0.0}$ & 47.4$_{\pm0.0}$ & 0.00$_{\pm0.00}$ & 0.0$_{\pm0.0}$ & 45.0$_{\pm0.0}$ & 0.00$_{\pm0.00}$ \\
  & $\mathbb{D}^2$                   & 0.0$_{\pm0.0}$ & 47.4$_{\pm0.0}$ & 0.00$_{\pm0.00}$ & 43.1$_{\pm3.3}$ & 56.6$_{\pm1.8}$ & 0.36$_{\pm0.03}$ \\
  & FA                               & 44.5$_{\pm3.3}$ & 61.0$_{\pm1.3}$ & 0.41$_{\pm0.03}$ & 29.2$_{\pm2.2}$ & 38.3$_{\pm4.1}$ & 0.15$_{\pm0.03}$ \\
  & FD                               & 43.5$_{\pm8.6}$ & 60.9$_{\pm3.8}$ & 0.40$_{\pm0.09}$ & \textbf{46.4$_{\pm2.4}$} & \textbf{60.0$_{\pm1.2}$} & \textbf{0.41$_{\pm0.03}$} \\
  & K-Center                         & \textbf{45.5$_{\pm4.3}$} & \textbf{61.6$_{\pm1.7}$} & \textbf{0.42$_{\pm0.04}$} & 40.0$_{\pm3.6}$ & 51.7$_{\pm3.0}$ & 0.30$_{\pm0.05}$ \\
  & LC                               & 39.4$_{\pm2.5}$ & 57.2$_{\pm1.9}$ & 0.35$_{\pm0.03}$ & 21.9$_{\pm0.4}$ & 19.1$_{\pm1.8}$ & 0.05$_{\pm0.01}$ \\
\rowcolor[HTML]{EEB3AA}
  & \ours-G (Ours)                     & 35.7$_{\pm5.1}$ & 58.4$_{\pm1.9}$ & 0.33$_{\pm0.05}$ & 44.6$_{\pm3.1}$ & 58.1$_{\pm2.0}$ & 0.38$_{\pm0.04}$ \\
\bottomrule
\end{tabular}
}
\end{table}

\noindent \textbf{\ours-G provides the largest gains at extreme budgets.}
At 1\% pruning, where random selection degrades substantially (F1 $23.8$ on both datasets with Tiny-CD and $13.9$/$6.3$ on \levir/\redd with BIT), \ours-G starts from a small random seed and iteratively adds high-quality samples.
On Tiny-CD, \ours-G achieves the best mIoU on \levir ($60.5$, compared with $60.4$ for K-Center and $54.5$ for random selection) and the best overall performance on \redd (F1 $53.2$, mIoU $63.7$), outperforming K-Center ($51.8$, $62.4$) and FD ($47.4$, $61.2$) by clear margins.
On BIT, \ours-G substantially recovers from the collapse of random selection (F1 $13.9 \rightarrow 35.7$ on \levir and $6.3 \rightarrow 44.6$ on \redd), although K-Center and FD retain an advantage on the \levir and \redd splits, respectively.
The improvement over random selection is most pronounced at 1\%, where the descriptor-guided construction directly addresses the distributional constraints identified in \autoref{sec:analysis_transition}. At this budget, even a single missing change-class example can eliminate positive supervision, and \ours-G's fidelity-aware scoring helps prevent such failures.

\begin{table}[t]
\centering
\footnotesize
\setlength{\tabcolsep}{3pt}
\caption{Ablation results (\%, except for $\kappa$) of removing each descriptor from sample scoring. CD Backbone: Tiny-CD.}
\label{tab:ablation_component}
\resizebox{\linewidth}{!}{%
\begin{tabular}{cl ccc | ccc}
\toprule
& & \multicolumn{3}{c}{\levir} &
    \multicolumn{3}{c}{\redd} \\
\cmidrule(lr){3-5} \cmidrule(lr){6-8}
Ratio & Pruner & F1 & mIoU & $\kappa$ & F1 & mIoU & $\kappa$ \\
\midrule
\multirow{5}{*}{10\%}
  & Full                     & \textbf{68.6}$_{\pm1.3}$ & \textbf{74.5}$_{\pm0.8}$ & \textbf{0.67}$_{\pm0.01}$ & \textbf{56.0}$_{\pm1.5}$ & \textbf{65.8}$_{\pm0.8}$ & \textbf{0.53}$_{\pm0.01}$ \\
  & \quad - I1                      & 66.6$_{\pm0.6}$ & 73.3$_{\pm0.3}$ & 0.65$_{\pm0.01}$ & 52.9$_{\pm1.9}$ & 64.2$_{\pm0.9}$ & 0.49$_{\pm0.02}$ \\
  & \quad - L1                      & 67.2$_{\pm2.2}$ & 73.7$_{\pm1.2}$ & 0.66$_{\pm0.02}$ & 55.5$_{\pm1.6}$ & 65.5$_{\pm0.8}$ & 0.52$_{\pm0.02}$ \\
  & \quad - F1                      & \textbf{68.6}$_{\pm0.9}$ & 74.4$_{\pm0.6}$ & \textbf{0.67}$_{\pm0.01}$ & 54.9$_{\pm2.1}$ & 65.3$_{\pm1.1}$ & 0.51$_{\pm0.02}$ \\
  & \quad - F2                      & 66.4$_{\pm1.4}$ & 73.2$_{\pm0.8}$ & 0.65$_{\pm0.01}$ & 55.0$_{\pm3.7}$ & 65.4$_{\pm1.9}$ & 0.52$_{\pm0.04}$ \\

\midrule
\multirow{5}{*}{5\%}
  & Full                            & \textbf{64.7}$_{\pm1.8}$ & \textbf{72.3}$_{\pm1.0}$ & \textbf{0.63}$_{\pm0.02}$ & 56.4$_{\pm2.2}$ & 65.9$_{\pm1.1}$ & 0.53$_{\pm0.02}$ \\
  & \quad - I1                      & 62.0$_{\pm2.5}$ & 70.7$_{\pm1.4}$ & 0.60$_{\pm0.03}$ & 56.3$_{\pm1.0}$ & 65.9$_{\pm0.5}$ & 0.53$_{\pm0.01}$ \\
  & \quad - L1                      & 62.7$_{\pm1.1}$ & 71.1$_{\pm0.6}$ & 0.61$_{\pm0.01}$ & 58.7$_{\pm0.1}$ & 67.0$_{\pm0.1}$ & 0.55$_{\pm0.00}$ \\
  & \quad - F1                      & 63.1$_{\pm1.3}$ & 71.4$_{\pm0.7}$ & 0.61$_{\pm0.01}$ & 58.2$_{\pm1.1}$ & 66.8$_{\pm0.5}$ & 0.54$_{\pm0.01}$ \\
  & \quad - F2                      & 62.7$_{\pm3.3}$ & 71.1$_{\pm1.8}$ & 0.61$_{\pm0.03}$ & \textbf{59.1}$_{\pm0.5}$ & \textbf{67.2}$_{\pm0.2}$ & \textbf{0.55}$_{\pm0.00}$ \\

  \midrule
\multirow{5}{*}{1\%}
  & Full                     & 41.1$_{\pm2.6}$ & 60.5$_{\pm1.1}$ & 0.39$_{\pm0.03}$ & \textbf{53.2}$_{\pm1.7}$ & \textbf{63.7}$_{\pm0.5}$ & \textbf{0.49}$_{\pm0.01}$ \\  
  & \quad - I1                      & 27.6$_{\pm19.9}$ & 55.6$_{\pm6.2}$ & 0.25$_{\pm0.18}$ & \textbf{53.2}$_{\pm1.2}$ & \textbf{63.7}$_{\pm0.5}$ & \textbf{0.49}$_{\pm0.01}$ \\
  & \quad - L1                      & \textbf{43.9}$_{\pm8.1}$ & \textbf{61.1}$_{\pm3.5}$ & \textbf{0.41}$_{\pm0.08}$ & 49.9$_{\pm1.8}$ & 62.0$_{\pm0.6}$ & 0.45$_{\pm0.01}$ \\
  & \quad - F1                      & 14.6$_{\pm20.7}$ & 52.2$_{\pm6.7}$ & 0.14$_{\pm0.20}$ & 52.1$_{\pm2.7}$ & 63.1$_{\pm1.1}$ & 0.47$_{\pm0.02}$ \\
  & \quad - F2                      & 14.6$_{\pm20.7}$ & 52.2$_{\pm6.7}$ & 0.14$_{\pm0.20}$ & 52.3$_{\pm0.9}$ & 63.2$_{\pm0.4}$ & 0.48$_{\pm0.01}$ \\

\bottomrule

\end{tabular}
}
    \vspace{-10pt}

\end{table}

\subsection{Ablation: Component Contributions}
\label{sec:ablation}
To isolate the contribution of each descriptor score, we conduct ablations with one of I1, L1, F1, and F2 removed during pruning.
We run at 10\%, 5\%, and 1\% pruning ratios on both \levir and \redd using Tiny-CD, averaged over three seeds. Results are reported in \autoref{tab:ablation_component}.

\noindent \textbf{At moderate budgets, I1 and F2 are the most impactful components.}
At 10\%, removing I1 causes the largest drop on both datasets (F1: $-2.0$ on \levir, $-3.1$ on \redd), followed by F2 ($-2.2$, $-1.0$).
Removing L1 has a moderate effect ($-1.4$ on \levir, $-0.5$ on \redd), while F1 is nearly redundant on \levir at this budget.
At 5\% on \levir, all four terms contribute, with I1 causing the largest drop ($-2.7$ F1), mirroring I1's leading univariate $R^2$ of $0.654$ in the regression analysis (\autoref{tab:descriptors}).
On \redd at 5\%, however, removing L1, F1, or F2 slightly \emph{improves} performance (e.g., removing F2: F1 $59.1$ vs.\ Full $56.4$).
Although L1 dominates the regression on \redd ($R^2 = 0.889$, RF $= 1.781$), this was measured across \emph{random} subsets with unconstrained change-ratio variation.
\ours-P starts from a random pool that already approximates the full-data distribution, so L1's corrective role is largely pre-satisfied, and the remaining scoring terms can over-constrain sample selection on a dataset with limited visual diversity.

\begin{figure}[t]
    \centering

    \begin{subfigure}[b]{0.49\linewidth}
        \centering
        \includegraphics[width=\linewidth]{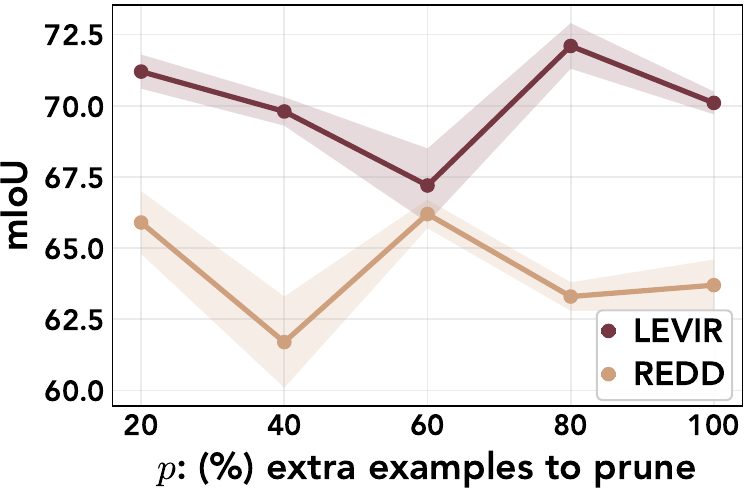}
        \caption{Smaller $p$ is competitive and more efficient ($r=5\%$).}
        \label{fig:abs_p}
    \end{subfigure}
    \begin{subfigure}[b]{0.49\linewidth}
        \centering
        \includegraphics[width=\linewidth]{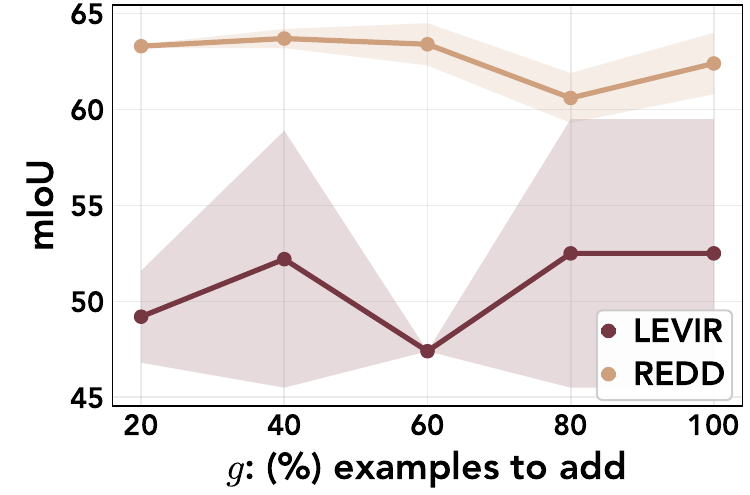}
        \caption{Larger $g$ is more efficient and stable in performance ($r=1\%$).}
        \label{fig:abs_g}
    \end{subfigure}
    \caption{Sensitivity of \ours-P to $p$ and \ours-G to $g$.}
    \label{fig:combined}
    \vspace{-8pt}
\end{figure}

\noindent \textbf{At extreme budgets, diversity and feature descriptors become critical on \levir.}
At 1\% on \levir, removing I1 degrades F1 from $41.1$ to $27.6$ ($\pm19.9$), and removing F1 or F2 causes near-complete collapse (F1 $14.6$, $\pm20.7$), indicating that diversity and feature-based scoring are essential when only 4--5 examples must span the full range of urban scene types.
In contrast, removing L1 at 1\% on \levir yields a marginally \emph{higher} mean F1 ($43.9$ vs.\ $41.1$), though with substantially larger variance ($\pm8.1$ vs.\ $\pm2.6$).
This is consistent with the regression analysis, where L1's univariate $R^2$ collapses to $0.000$ at 1\% on \levir (\autoref{tab:descriptors}): with only 4-5 examples, change-ratio deviation becomes a poor linear predictor of performance, and enforcing it as a scoring term can prevent the inclusion of otherwise high-quality diverse examples, occasionally yielding a better coreset but at the cost of reliability.
On \redd at 1\%, the pattern is reversed: L1 is the only component whose removal meaningfully hurts (F1 $49.9$ vs.\ $53.2$), while I1 has no effect ($53.2$ in both cases).
This directly parallels the regression, where L1 retains $R^2 = 0.823$ and RF importance $1.722$ on \redd at 1\%, while I1's RF importance is negligible ($0.023$), reaffirming \redd as a distribution-dominated regime.

\noindent \textbf{Overall, the ablation results validate the regression-derived design.}
The dataset-dependent importance hierarchy observed in the ablation: I1 critical on visually heterogeneous \levir and L1 indispensable on distribution-sensitive \redd, closely mirrors the regression findings in \autoref{sec:analysis}.
The cases where ablation results diverge from the regression (e.g., L1's positive effect at 1\% on \levir, or the REDD over-constraining at 5\%) are explained by the fact that \ours operates on a pre-filtered pool rather than unconstrained random subsets, shifting the marginal value of each descriptor relative to the regression baseline.

\subsection{Sensitivity of \ours-P to $p$ and \ours-G to $g$}
\label{sec:ablation_pg}
To investigate the effects of changing $p$ and $g$ in the two optimization directions of \ours, we vary $p$ from 20\% to 100\% under $r=5\%$ and $g$ from 20\% to 100\% under $r=1\%$, both training for the Tiny-CD model. \autoref{fig:abs_p} shows that starting with too many examples to prune leads to slightly worse performance, suggesting that at moderate pruning ratio preserving the random structure is beneficial. \autoref{fig:abs_g} shows that starting with fewer examples to add later results in lower performance variance, indicating that preserving the random structure leads to more stable training. In practice, we choose $p=20\%$ and $g=40\%$, as this setting requires fewer iterations while maintaining competitive performance.

\section{Limitations and Future Work}
Our study focuses on binary change detection across two benchmarks with distinct land-cover characteristics. Extending and validating the proposed taxonomy of diversity-limited and distribution-shift settings across a broader range of datasets remains an important direction for future work. In addition, future work should investigate pruning under alternative training paradigms, including semi-supervised and self-supervised learning pipelines. Finally, learning dataset-specific descriptor weights from a small pilot regression, rather than relying on hand-tuned configurations, represents a promising direction toward fully automated coreset construction.

\section{Conclusion}
\label{sec:conclusion}
We present the first systematic study of coreset selection for remote sensing change detection (RS-CD), comprising a benchmark evaluation, a descriptor-based analysis, and a principled selection method.

Benchmarking six adapted pruning criteria on \levir and \redd with Tiny-CD and BIT across pruning ratios of 10\%, 5\%, and 1\%, we find that \emph{no existing method reliably outperforms random selection}. Loss-based criteria fail because prediction difficulty does not correspond to supervisory value in CD; label-complexity selection destroys distributional coverage; and feature-geometry methods (FD, K-Center) offer consistent gains only at extreme budgets. These failures are structural rather than architecture-dependent.

Descriptor analysis reveals a dataset-dependent priority hierarchy: on \levir, intra-set diversity (I1) dominates at moderate budgets, while on \redd, change distribution fidelity (L1) is near-sufficient. At 1\% budgets, importance of diversity depends more on the dataset, with label-feature consistency (F2) providing a complementary discriminating signal.

Guided by this analysis, we propose Fidelity-Diversity-Consistency (\ours), which enforces change distribution fidelity as a hard constraint while promoting diversity and consistency as soft objectives. \ours-P and \ours-G consistently outperform the already competitive random-selection baseline, particularly at extreme pruning ratios. 

Together, these findings reframe the central challenge of RS-CD coreset selection: the goal is not to identify the hardest samples, but to preserve the distributional properties that govern change-learning dynamics.

\bibliographystyle{plainnat}
\bibliography{reference}

\end{document}